\documentclass[runningheads]{llncs}

\usepackage[T1]{fontenc}
\usepackage{graphicx}
\usepackage{amsmath}
\usepackage{booktabs}
\usepackage{array}
\usepackage{float}
\usepackage{longtable}
\usepackage[utf8]{inputenc}
\usepackage{silence}
\usepackage{hyperref}
\usepackage{xcolor}
\usepackage{listings}
\usepackage{hyperref} 

\colorlet{punct}{red!60!black}
\definecolor{background}{HTML}{EEEEEE}
\definecolor{delim}{RGB}{20,105,176}
\colorlet{numb}{magenta!60!black}

\lstdefinelanguage{json}{
    basicstyle=\scriptsize\ttfamily,
    numbers=left,
    numberstyle=\scriptsize,
    stepnumber=1,
    numbersep=8pt,
    showstringspaces=false,
    breaklines=true,
    frame=lines,
    backgroundcolor=\color{background},
    extendedchars=true,
    literate=
     *{0}{{{\color{numb}0}}}{1}
      {1}{{{\color{numb}1}}}{1}
      {2}{{{\color{numb}2}}}{1}
      {3}{{{\color{numb}3}}}{1}
      {4}{{{\color{numb}4}}}{1}
      {5}{{{\color{numb}5}}}{1}
      {6}{{{\color{numb}6}}}{1}
      {7}{{{\color{numb}7}}}{1}
      {8}{{{\color{numb}8}}}{1}
      {9}{{{\color{numb}9}}}{1}
      {:}{{{\color{punct}{:}}}}{1}
      {,}{{{\color{punct}{,}}}}{1}
      {\{}{{{\color{delim}{\{}}}}{1}
      {\}}{{{\color{delim}{\}}}}}{1}
      {[}{{{\color{delim}{[}}}}{1}
      {]}{{{\color{delim}{]}}}}{1}
      {’}{{'}}1 
      {—}{---}1 
      {“}{{"}}1 
      {”}{{"}}1,
}
\providecommand{\doi}[1]{\url{#1}}

\begin{document}
%
\title{How well do Large Language Models Recognize Instructional Moves? Establishing Baselines for Foundation Models in Educational Discourse}



\titlerunning{Baselines for Foundation Models}
%
\author{Kirk Vanacore  \and René F. Kizilcec}

\authorrunning{K. Vanacore and R.F. Kizilcec}
%
 \institute{National Tutoring Observatory\\Cornell University}
 \maketitle              
\begin{abstract}
Large language models (LLMs) are increasingly adopted in educational technologies for a variety of tasks, from generating instructional materials and assisting with assessment design to tutoring. While prior work has investigated how models can be adapted or optimized for specific tasks, far less is known about how well LLMs perform at interpreting authentic educational scenarios without significant customization. As LLM-based systems become widely adopted by learners and educators in everyday academic contexts, understanding their out-of-the-box capabilities is increasingly important for setting expectations and benchmarking. We compared six LLMs to estimate their baseline performance on a simple but important task: classifying instructional moves in authentic classroom transcripts. We evaluated typical prompting methods: zero-shot, one-shot, and few-shot prompting. We found that while zero-shot performance was moderate, providing comprehensive examples (few-shot prompting) significantly improved performance for state-of-the-art models, with the strongest configuration reaching $\kappa = 0.58$ against expert-coded annotations. At the same time, improvements were neither uniform nor complete: performance varied considerably by instructional move, and higher recall frequently came at the cost of increased false positives. Overall, these findings indicate that foundation models demonstrate meaningful yet limited capacity to interpret instructional discourse, with prompt design helping to surface capability but not eliminating fundamental reliability constraints. 

\end{abstract}
\section{Introduction}

Research on generative artificial intelligence (GenAI) in education has advanced our understanding of how large language models (LLMs) can be harnessed to improve or extend existing educational technologies. For instance, extensive research has examined whether LLMs can be tailored for specific instructional tasks \cite{Zhang2024_GenAIReview}, such as generating feedback or producing learning materials \cite{Lee_2024,Meyer_2024}. Yet an equally pressing question concerns how well these general-purpose foundation models perform when used without task-specific tuning in authentic educational contexts. The widespread availability of LLMs is already reshaping how students and educators seek academic support: students report using AI for assignment completion~\cite{Turos2025AIHomework} and obtaining help while learning~\cite{Adams2024_AIhelp}. One recent study found that a large majority of university students now use ChatGPT for academic purposes, including concept explanation, problem-solving, and clarifying assignment expectations~\cite{Mohammad2025_useChatGPT}. Furthermore, there is growing evidence that teachers rely on AI to create educational content \cite{VanDenBerg_2023,DelValle_2025}, while many districts are not providing training on the use of AI in the classroom \cite{Diliberti_2025}. Collectively, these patterns suggest that openly available foundation models already function as \textit{de facto} educational tools regardless of whether they have been explicitly adapted or optimized for such tasks.

Researchers have developed benchmarks to evaluate the baseline performance of foundation models, including LLMs released by OpenAI, Anthropic, and Alphabet. These benchmarks elucidate how models perform \textit{out-of-the-box} on education-related tasks, rather than how they perform when used by researchers with carefully engineered prompts, curated examples, downstream verification methods, or even fine-tuning. These idiosyncratic customizations make it difficult to separate genuine model capability from scaffolded performance. Researchers have examined how AI models perform on educational assessments, understand problems, and identify students' misconceptions \cite{otero2025benchmark,yang2025mmtutorbench,macina2025mathtutorbench,Lelièvre_2025}. We add to the growing body of work on educational benchmarking of AI models by examining how current foundation models behave under minimal guidance when classifying instructional moves in real instructional transcripts, establishing an empirical baseline for future methodological and applied work. We find that leading foundation models (e.g., Claude 4.5 Opus) can achieve moderate reliability when provided with clear definitions and extensive examples (few-shot). However, the observed patterns of performance also suggest potential weaknesses in LLMs' current ability to evaluate the complex blend of social and cognitive discourse that underlies effective instruction without such support.

\subsection{Current Benchmarks AI Performance in Educational Contexts}

Recent efforts in educational AI benchmarking span a range of tasks, from pedagogical test performance to diagnosing specific algebraic misconceptions. For example, the Pedagogy Benchmark leaderboard\footnote{\url{https://benchmarks.ai-for-education.org/}} focuses on broader pedagogical alignment, particularly regarding cross-domain pedagogical knowledge assessed through tests on pedagogical theory and practice \cite{Lelièvre_2025}. Many foundation models perform well, answering 80\%--90\% of questions correctly.

Other benchmarks show that foundation model performance varies on more complex tasks. Otero et al.~\cite{otero2025benchmark} provide a benchmark that focuses on diagnostic capability, testing whether models can identify 55 distinct algebraic misconceptions (such as variable cancellation errors) rather than simply generating a correct solution. They found that current AI models exhibit varying performance in correctly identifying misconceptions, depending on the type of math problem (e.g., lower performance on ratio problems and higher performance on number operations). Yang et al.~\cite{yang2025mmtutorbench} extended this work by introducing MMTutorBench, which evaluates multimodal tutoring capabilities. Their findings reveal a critical gap: while models may correctly recognize text, they struggle to integrate visual student work (like handwritten equations) into a coherent tutoring strategy, often failing to distinguish between a student's insight error and a mere execution error. In the domain of open-ended dialogue, Macina et al.~\cite{macina2025mathtutorbench} highlight an "expertise-pedagogy trade-off." Their work on MathTutorBench shows that models optimized for high-level mathematical problem-solving often exhibit lower pedagogical quality, struggling to perform "Socratic" moves such as scaffolding a hint without revealing the answer. 

\subsection{AI Identification of Instructional Moves}

Annotation is a common benchmarking task for AI models, which may be particularly important for assessing whether these models can operate in educational contexts similarly to human experts. Recent research examines how LLMs can streamline qualitative analysis of educational data, including workflows specifically designed for student and teacher discourse. For human-in-the-loop workflows, Henkel and Hills~\cite{Henkel2023_ComparativeJudgment} demonstrate that AI-assisted comparative judgment improves reliability when scoring open-ended student responses. Similarly, Barany et al. and Zambrano et al. find that LLMs function as "critical partners" in quantitative ethnography, helping researchers generate, refine, and consolidate codebooks for analyzing complex learning interactions~\cite{Barany2024_CodebookDev,Zambrano2023_nCoderChatGPT}.

When evaluating LLMs as an independent labeler of instructional content, performance is promising but highly context-dependent. Liu et al.~\cite{Liu2024_PotentialLimits,Liu2025_GPT4Better} report that GPT-4 can match human reliability in coding virtual tutoring transcripts, though it struggles with underrepresented constructs. In the context of whole-class discussion, Long et al.~\cite{Long2024_ClassroomDialogue} find that while LLMs achieve moderate agreement with experts, they exhibit systematic weaknesses in identifying nuanced instructional moves. Methodological challenges remain a core focus: Nahum et al. warn that errors in human "gold standard" data can distort model evaluation~\cite{Nahum2025_LabelErrors}, while He et al.~\cite{He2025_PromptingDark} note that without high-quality gold labels, even experts struggle to improve model performance through iterative prompting.

Finally, while few-shot prompting (providing examples within the prompt) generally improves performance in educational coding tasks~\cite{Liu2025_GPT4Better}, it is not a panacea \cite{Geathers2025_OSCEBenchmark}. Studies on classroom dialogue indicate that adding examples can sometimes degrade performance if the examples are not perfectly aligned with the specific pedagogical context~\cite{Long2024_ClassroomDialogue}, suggesting that rigorous, construct-specific evaluation is essential for deploying AI in educational research.

\section{Current Study}

Despite rapid advances in AI-assisted educational analytics, less is known about the \textit{baseline} capability of foundation models to interpret complex pedagogical discourse without extensive prompting, fine-tuning, or verification scaffolds. Prior work has largely focused on optimizing model performance rather than establishing how well unadjusted models understand instructional interactions. This study addresses that gap by establishing baseline performance estimates for several state-of-the-art foundation models on a core task in instructional discourse interpretation.

We focus on a well-validated and theoretically consequential construct for instructional quality: \textit{Talk Moves}, grounded in accountable talk theory \cite{Michaels2008AccountableTalk,Suresh_2022_TalkMoves}. These moves represent a set of discourse strategies---such as pressing for reasoning, probing student thinking, or revoicing contributions---that capture how teachers elicit, respond to, and build on student ideas. Identifying \textit{Talk Moves} requires sensitivity to conversational context, speaker intention, and pedagogical purpose, making this construct an especially demanding test of foundation model interpretive ability.

To evaluate model performance, we benchmark six leading models against expert-annotated classroom transcripts. We assess their abilities under minimal guidance, systematically varying prompting strategies (e.g., zero-shot, one-shot, and few-shot prompts) to estimate how different levels of support affect baseline accuracy. Performance is evaluated using both global agreement metrics and construct-specific analyses to identify patterns of strength and weakness across move types.

This study investigates three research questions:

\begin{description}
  \item[\textbf{RQ1:}] How do state-of-the-art LLMs perform on identifying \textit{Talk Moves} in authentic classroom discourse under minimal guidance?
  \item[\textbf{RQ2:}] Which specific discourse constructs pose persistent challenges for LLMs, and where do LLMs exhibit comparatively stronger interpretive capability?
  \item[\textbf{RQ3:}] How do prompting methods affect model performance? Specifically, does giving the LLM instructions and examples that human coders receive significantly improve performance?

\end{description}

\section{Method}

\subsection{Foundation Models}
To assess the relationship between general-purpose capability and specific educational performance, we selected models based on their ranking on the \textit{AI-for-Education.org}\footnote{\url{https://benchmarks.ai-for-education.org/}} Pedagogy Benchmark leaderboard (as of December 2025). We included the top-ranking models available to us to ensure a representative sample of state-of-the-art performance. Table \ref{tab:model_selection} details six of the top models from the leaderboard, which we evaluated in this study.

\begin{table}[H]
\centering
\caption{Six Top Performing Foundation Models on AI-For-Education Benchmark}
\label{tab:model_selection}
\begin{tabular}{l l c}
\toprule
\textbf{Model} & \textbf{Model ID} & \textbf{CDPK Performance} \\
\midrule
Gemini-3 Pro & \texttt{google.gemini-3-pro-preview} & 91\% \\
Gemini 2.5 Pro & \texttt{google.gemini-2.5-pro} & 89\% \\
GPT-5 & \texttt{openai.gpt-5} & 88\% \\
o3 & \texttt{openai.o3} & 88\% \\
Claude 4.5 Opus & \texttt{anthropic.claude-4.5-opus} & 87\% \\
Claude 4.5 Sonnet & \texttt{anthropic.claude-4.5-sonnet} & 82\% \\
\bottomrule
\end{tabular}
\end{table}

\subsection{Experimental Design}

We employed a factorial design crossing the selected foundation models with four prompting strategies:

\begin{itemize}
    \item \textit{Zero-Shot}: Task description and code definitions only, with no examples.
    \item \textit{One-Shot}: Task description plus a single annotated example for each talk move.
    \item \textit{Few-Shot (Three Examples)}: Task description plus three annotated examples for each talk move.
    \item \textit{Few-Shot (All Examples)}: Task description, code definitions, and all annotated examples provided in the official \textit{Talk Moves} documentation manual.
\end{itemize}

All prompts (see Appendix B) were adapted directly from the Coding Manual\footnote{\url{https://github.com/SumnerLab/TalkMoves/blob/main/Coding\%20Manual.pdf}} used by the human annotators. Our goal was to mirror the information available to human coders as closely as possible, including definitions, decision rules, and examples. The Few-Shot prompt (All Examples), therefore, incorporated the full set of codebook examples, ranging from 3 to 13 per move. Although providing such an uneven and extensive set of examples is atypical in few-shot prompting, we adopted this structure to align the prompt with what expert human annotators receive during training.

All data and prompts were passed to the models in JSON format, with additional instructions to enforce a consistent output structure across conditions. Dialogue was submitted to the LLMs in chunks of 30 discourse lines, each containing a single utterance, to provide sufficient---but not overwhelming---context for interpreting instructional interactions. Each model–prompt combination was applied to the entire dataset, yielding 12 experimental conditions. The code to run these experiments is available on the National Tutoring Observatory Open-source GitHub Repository\footnote{\url{https://github.com/National-Tutoring-Observatory/Baseline_Performance_Pipeline.git}}.

\subsection{Sample \& Data}
We used the \textit{TalkMoves Dataset} of classroom transcripts that were previously annotated and validated by expert educators to create ground truth labels \cite{Suresh_2022_TalkMoves}. The subset used for this analysis consists of 63 transcripts spanning a diverse range of grade levels. The dataset comprises mathematics classroom transcripts drawn from authentic K–12 instructional settings, including whole-class discussions, small-group problem-solving, and online lesson contexts. Each transcript was human-transcribed, segmented by speaker, and annotated at the utterance level with six pedagogically grounded \textit{Talk Moves} reflecting the Accountable Talk framework. This coding scheme includes the following moves:

\begin{itemize}
    \item \textit{Keeping Everyone Together}: Ensuring all students follow the discussion
    \item \textit{Getting Students to Relate to Another's Ideas}: Promoting peer engagement
    \item \textit{Restating}: Repeating student contributions verbatim
    \item \textit{Revoicing}: Rephrasing student ideas in academic language
    \item \textit{Pressing for Accuracy}: Requesting precise or correct responses
    \item \textit{Pressing for Reasoning}: Eliciting student explanations
    \item \textit{None}: No identifiable talk move present
\end{itemize}

The ground truth annotations served as the gold standard for evaluating model performance. These annotations were conducted by human experts in math instruction and accountable talk theory. These annotators achieved a high inter-rater reliability (Cohen's $\kappa > 0.90$) \cite{Suresh_2022_TalkMoves}. For our analysis, the transcripts were chunked into segments of up to 20 utterances to accommodate model context windows while preserving conversational coherence.

Because running 4 prompts per model is costly, we ran this experiment on a representative sample of the original data. We used a proportional stratified random sampling method to select 800 target utterances that match the ground truth distribution of Talk Moves, which were then expanded to include their surrounding conversation (20 turns prior, 1 turn post) and merged into contiguous blocks. This process yielded 467 unique chunks containing 13,239 utterances. Due to the chunking process, more utterances were included in the sample than the originally targeted 800; approximately 79.5\% (10,522) of these utterances were teacher turns with valid codes, while the remaining 20.5\% provided necessary conversational context (student turns or unlabeled teacher turns).

\section{Results}

\subsection{Baseline Performance of Foundation Models}

To address our first research question regarding the baseline capability of foundation models, we examined the overall agreement between model and expert human annotations across all conditions and models. Table \ref{tab:model_performance} presents the Precision, Recall, F1 score, and Cohen's $\kappa$ for each model-prompt combination.

\begin{table}[H]
\centering
\small
\setlength\tabcolsep{4pt}
\caption{Overall Model Performance. \textbf{Bold} values indicate best performance within each prompting strategy; \textit{italic} values indicate worst performance. Asterisks indicate significant difference from Zero Shot baseline: * p $<$ .05, ** p $<$ .01, *** p $<$ .001.}
\label{tab:model_performance}
\begin{tabular}{l>{ \centering \arraybackslash}p{1.8cm}>{ \centering \arraybackslash}p{1.8cm}>{ \centering \arraybackslash}p{1.8cm}>{ \centering \arraybackslash}p{1.8cm}}
  \toprule
Model & Precision & Recall & F1 & Kappa \\
  \midrule
\addlinespace[0.5em] \multicolumn{5}{l}{\textbf{Zero Shot}} \\
Gemini 3 Pro & 0.50 & 0.63 & \begin{tabular}[c]{@{}c@{}}0.51\\\scriptsize{(0.50-0.52)} \end{tabular} & \begin{tabular}[c]{@{}c@{}}0.48\\\scriptsize{(0.46-0.49)} \end{tabular} \\
Claude 4.5 Opus & 0.49 & \textbf{0.67} & \begin{tabular}[c]{@{}c@{}}0.52\\\scriptsize{(0.51-0.53)} \end{tabular} & \begin{tabular}[c]{@{}c@{}}\textbf{0.48}\\\scriptsize{(0.47-0.49)} \end{tabular} \\
Claude 4.5 Sonnet & \textit{0.44} & 0.60 & \begin{tabular}[c]{@{}c@{}}\textit{0.45}\\\scriptsize{(0.44-0.46)} \end{tabular} & \begin{tabular}[c]{@{}c@{}}0.40\\\scriptsize{(0.39-0.42)} \end{tabular} \\
Gemini 2.5 & 0.47 & \textit{0.59} & \begin{tabular}[c]{@{}c@{}}0.47\\\scriptsize{(0.46-0.48)} \end{tabular} & \begin{tabular}[c]{@{}c@{}}0.38\\\scriptsize{(0.37-0.40)} \end{tabular} \\
GPT-5 & \textbf{0.50} & 0.66 & \begin{tabular}[c]{@{}c@{}}\textbf{0.53}\\\scriptsize{(0.52-0.54)} \end{tabular} & \begin{tabular}[c]{@{}c@{}}0.47\\\scriptsize{(0.46-0.49)} \end{tabular} \\
o3 & 0.48 & 0.65 & \begin{tabular}[c]{@{}c@{}}0.51\\\scriptsize{(0.50-0.52)} \end{tabular} & \begin{tabular}[c]{@{}c@{}}0.46\\\scriptsize{(0.45-0.48)} \end{tabular} \\
\addlinespace[0.5em] \multicolumn{5}{l}{\textbf{One Shot}} \\
Gemini 3 Pro & 0.50 & 0.63 & \begin{tabular}[c]{@{}c@{}}0.50\\\scriptsize{(0.49-0.51)} \end{tabular} & \begin{tabular}[c]{@{}c@{}}0.48\\\scriptsize{(0.46-0.49)} \end{tabular} \\
Claude 4.5 Opus & \textbf{0.51} & \textbf{0.67} & \begin{tabular}[c]{@{}c@{}}\textbf{0.53}\\\scriptsize{(0.52-0.54)} \end{tabular} & \begin{tabular}[c]{@{}c@{}}\textbf{0.49}\\\scriptsize{(0.47-0.50)} \end{tabular} \\
Claude 4.5 Sonnet & \textit{0.48} & \textit{0.60} & \begin{tabular}[c]{@{}c@{}}\textit{0.47}\\\scriptsize{(0.46-0.48)} \end{tabular} & \begin{tabular}[c]{@{}c@{}}0.44\\\scriptsize{(0.42-0.45)} \end{tabular} \\
Gemini 2.5 & 0.50 & 0.61 & \begin{tabular}[c]{@{}c@{}}0.48\\\scriptsize{(0.47-0.49)} \end{tabular} & \begin{tabular}[c]{@{}c@{}}0.45\\\scriptsize{(0.43-0.46)} \end{tabular} \\
GPT-5 & 0.51 & 0.66 & \begin{tabular}[c]{@{}c@{}}0.53\\\scriptsize{(0.52-0.54)} \end{tabular} & \begin{tabular}[c]{@{}c@{}}0.48\\\scriptsize{(0.47-0.50)} \end{tabular} \\
o3 & 0.49 & 0.64 & \begin{tabular}[c]{@{}c@{}}0.50\\\scriptsize{(0.49-0.51)} \end{tabular} & \begin{tabular}[c]{@{}c@{}}0.47\\\scriptsize{(0.46-0.49)} \end{tabular} \\
\addlinespace[0.5em] \multicolumn{5}{l}{\textbf{Few Shot (Three Examples)}} \\
Gemini 3 Pro & 0.53 & 0.68 & \begin{tabular}[c]{@{}c@{}}0.56\\\scriptsize{(0.55-0.57)} \end{tabular} & \begin{tabular}[c]{@{}c@{}}0.53\\\scriptsize{(0.51-0.54)} \end{tabular} \\
Claude 4.5 Opus & \textbf{0.54} & \textbf{0.70} & \begin{tabular}[c]{@{}c@{}}\textbf{0.57}\\\scriptsize{(0.56-0.58)} \end{tabular} & \begin{tabular}[c]{@{}c@{}}\textbf{0.54}\\\scriptsize{(0.53-0.56)} \end{tabular} \\
Claude 4.5 Sonnet & 0.51 & \textit{0.65} & \begin{tabular}[c]{@{}c@{}}\textit{0.52}\\\scriptsize{(0.51-0.53)} \end{tabular} & \begin{tabular}[c]{@{}c@{}}0.49\\\scriptsize{(0.47-0.50)} \end{tabular} \\
Gemini 2.5 & 0.52 & 0.67 & \begin{tabular}[c]{@{}c@{}}0.53\\\scriptsize{(0.52-0.54)} \end{tabular} & \begin{tabular}[c]{@{}c@{}}0.50\\\scriptsize{(0.48-0.51)} \end{tabular} \\
GPT-5 & 0.53 & 0.70 & \begin{tabular}[c]{@{}c@{}}0.56\\\scriptsize{(0.55-0.57)} \end{tabular} & \begin{tabular}[c]{@{}c@{}}0.51\\\scriptsize{(0.50-0.52)} \end{tabular} \\
o3 & \textit{0.48} & 0.67 & \begin{tabular}[c]{@{}c@{}}0.52\\\scriptsize{(0.51-0.53)} \end{tabular} & \begin{tabular}[c]{@{}c@{}}0.48\\\scriptsize{(0.46-0.49)} \end{tabular} \\
\addlinespace[0.5em] \multicolumn{5}{l}{\textbf{Few Shot (All Examples)}} \\
Gemini 3 Pro & 0.54 & 0.70 & \begin{tabular}[c]{@{}c@{}}0.59\\\scriptsize{(0.58-0.60)} \end{tabular} & \begin{tabular}[c]{@{}c@{}}0.54\\\scriptsize{(0.52-0.55)} \end{tabular} \\
Claude 4.5 Opus & \textbf{0.56} & \textbf{0.75} & \begin{tabular}[c]{@{}c@{}}\textbf{0.61}\\\scriptsize{(0.60-0.62)} \end{tabular} & \begin{tabular}[c]{@{}c@{}}\textbf{0.58}\\\scriptsize{(0.57-0.60)} \end{tabular} \\
Claude 4.5 Sonnet & 0.51 & 0.69 & \begin{tabular}[c]{@{}c@{}}0.55\\\scriptsize{(0.54-0.56)} \end{tabular} & \begin{tabular}[c]{@{}c@{}}0.52\\\scriptsize{(0.51-0.53)} \end{tabular} \\
Gemini 2.5 & 0.54 & 0.72 & \begin{tabular}[c]{@{}c@{}}0.59\\\scriptsize{(0.58-0.60)} \end{tabular} & \begin{tabular}[c]{@{}c@{}}0.57\\\scriptsize{(0.56-0.58)} \end{tabular} \\
GPT-5 & 0.54 & 0.72 & \begin{tabular}[c]{@{}c@{}}0.59\\\scriptsize{(0.58-0.60)} \end{tabular} & \begin{tabular}[c]{@{}c@{}}0.49\\\scriptsize{(0.48-0.51)} \end{tabular} \\
o3 & \textit{0.49} & \textit{0.68} & \begin{tabular}[c]{@{}c@{}}\textit{0.53}\\\scriptsize{(0.52-0.54)} \end{tabular} & \begin{tabular}[c]{@{}c@{}}0.45\\\scriptsize{(0.44-0.46)} \end{tabular} \\
   \bottomrule
\end{tabular}
\end{table}

Overall, foundation models demonstrated moderate classification accuracy, with Cohen's $\kappa$ ranging from 0.38 to 0.58 across all model-prompt combinations. In the baseline Zero Shot condition, performance was relatively consistent, with Claude 4.5 Opus and Gemini 3 Pro achieving the highest baseline agreement ($\kappa \approx 0.48$), followed closely by GPT-5 ($\kappa \approx 0.47$) and o3 ($\kappa \approx 0.46$). Claude 4.5 Sonnet and Gemini 2.5 Pro exhibited slightly lower baseline agreement ($\kappa = 0.40$ and $0.38$ respectively). Notably, the range of performance in the Zero Shot condition was relatively narrow, highlighting a consistent performance ceiling for current models when operating without examples.

Model performance for classifying Talk Moves was also moderate. Notably, all models displayed higher recall than precision, suggesting that while models were able to identify many true instances, they also tended to overproduce labels, resulting in a relatively high rate of false positives. Even the highest-performing model, Claude 4.5 Opus, achieved a precision of only 0.56, indicating that nearly half of the moves it identified were actually false positives, despite its ability to capture a high percentage of true instances (Recall = 0.75).

\subsection{Construct-Specific Performance}

\begin{figure}[H]
    \centering
    \includegraphics[width=\textwidth]{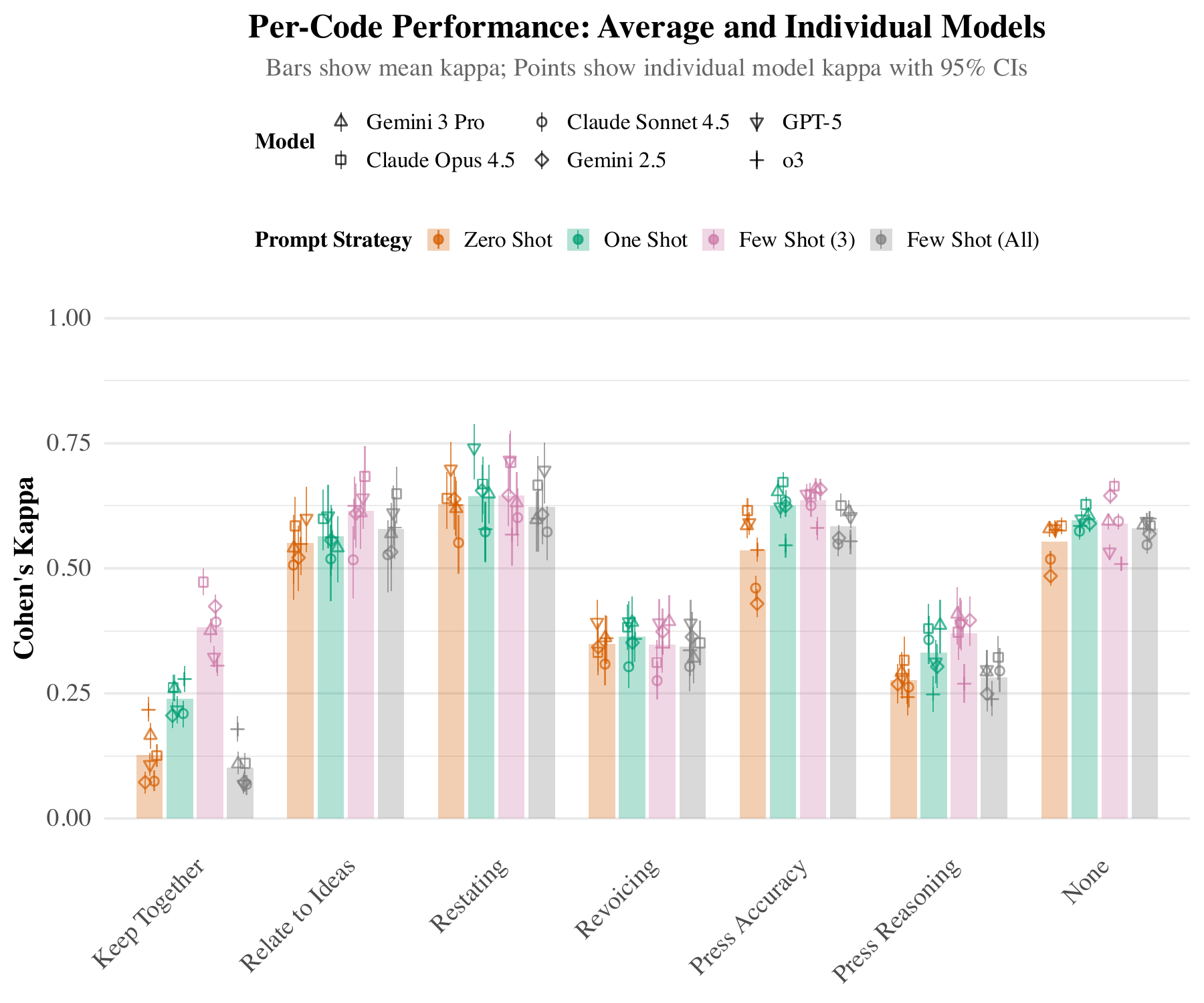}
    \caption{Per-Code Cohen's $\kappa$ by Prompt Strategy and Model. Bars represent mean $\kappa$; points show individual model performance with 95\% confidence intervals.}
    \label{fig:per_code_performance}
\end{figure}

Next, to address our second research question, we examined which specific discourse constructs posed the greatest challenges for LLMs. Figure \ref{fig:per_code_performance} displays per-code $\kappa$, revealing substantial variation across different types of \textit{Talk Moves}. (Tables with the full results can be found in Appendix \ref{App_precode}.)

Models reached moderate to substantial reliability on \textit{Getting Students to Relate to Another's Ideas}, \textit{Restating}, and \textit{Pressing for Accuracy}, whereas reliability mostly remained poor or fair for \textit{Keeping Everyone Together}, \textit{Revoicing}, and \textit{Pressing for Reasoning}. One possible explanation is that the models exhibit comparatively stronger interpretive capability on more explicit, structurally constrained discourse moves. In contrast, they struggled with moves that require sensitivity to social dynamics and metacognitive context. For example, it is comparatively easier to determine whether a teacher is prompting students to respond to one another's ideas than to detect the more subtle behavior of maintaining collective classroom orientation. Likewise, \textit{Restating} is inherently explicit, whereas \textit{Revoicing} requires inferring how an instructor reshapes a student’s contribution to advance shared understanding---a task that appears to challenge current foundation models.

\subsection{Effects of Prompting Strategies}

Addressing the third research question, we find substantial heterogeneity in how models respond to prompting strategies (Table \ref{tab:model_performance}). Generally, examples improved performance, though the magnitude of this benefit varied substantially by model. Claude 4.5 Opus demonstrated the most profound improvement, rising from a baseline $\kappa$ of 0.48 (Zero Shot) to 0.58 in the Few Shot (All Examples) condition. Similarly, Gemini 2.5 Pro showed a large gain, improving from 0.38 to 0.57. These models appear to effectively leverage the extensive definition and example context provided in the full prompt. In contrast, GPT-5 and o3 showed minimal sensitivity to prompting strategies, maintaining relatively stable performance across conditions. Gemini 3 Pro showed moderate improvement (0.48 to 0.54). 

These findings suggest that while some models (Opus, Gemini 2.5) possess a latent capacity for higher performance that can be unlocked through prompt engineering, others (GPT-5, o3) may be limited by their zero-shot baseline or less responsive to in-context learning for this specific task.

\begin{figure}[H]
    \centering
    \includegraphics[width=\textwidth]{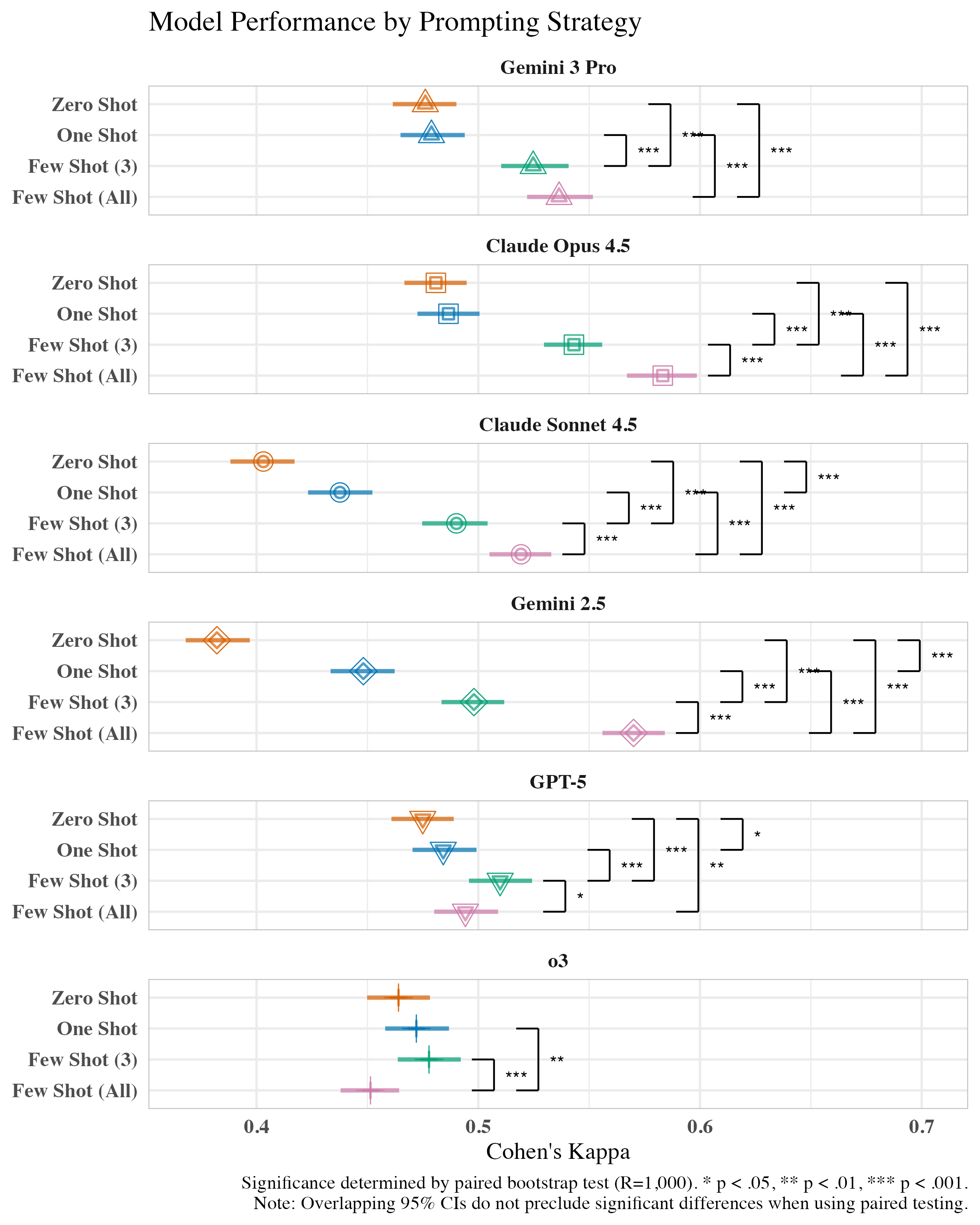}
    \caption{Kappa scores for each model across zero-shot, one-shot, and few-shot prompting conditions. Error bars represent 95\% bootstrapped confidence intervals. Brackets indicate statistically significant pairwise differences determined by paired bootstrap test ($R=1,000$). Note: Overlapping 95\% CIs do not preclude significant differences when using paired testing.}
    \label{fig:prompt_diffs}
\end{figure}

\section{Discussion}

This study set out to establish a baseline for how well current foundation models can interpret instructional discourse without fine-tuning, retrieval, or multi-stage verification pipelines. Across models and prompting conditions, we found that unadjusted foundation models exhibit moderate agreement ($\kappa = 0.38$--$0.58$) with expert human annotations. The strongest model–prompt configuration (Claude 4.5 Opus with the few-shot all examples prompt, $\kappa = 0.58$) achieved agreement close to the standard threshold for substantial inter-coder agreement \cite{LandisKoch1977}, approaching the reliability of expert human coders in some categories. However, even this model produced a high false positive rate. This represents a significant advance over earlier models. However, variability remains, and no model achieved perfect alignment "out of the box." Comparison of these findings provides some empirical evidence that, despite impressive general linguistic competence, today’s foundation models require careful prompting and model selection to interpret fine-grained pedagogical discourse reliably.

\subsection{Baseline Competence is Uneven}

A central contribution of this study is demonstrating that the "out-of-the-box" interpretive ability of leading foundation models remains highly uneven across instructional constructs. Models performed best on linguistically explicit and structurally constrained moves such as \textit{Restating} and \textit{Pressing for Accuracy}, which often rely on surface-level repetition or direct interrogative forms. In contrast, models consistently struggled with discourse moves that require sensitivity to classroom norms, collective attention, and inter-student relationships---most notably \textit{Keeping Everyone Together} and, under some conditions, \textit{Getting Students to Relate to Another's Ideas}. These findings align with prior work showing that LLMs perform well on pattern-detectable linguistic structures but degrade when social, pragmatic, or metacognitive inferences are required \cite{Long2024_ClassroomDialogue}. 

Critically, these weaknesses emerged even though all evaluated models rank highly on general-purpose pedagogy benchmarks (e.g., \hyperlink{https://benchmarks.ai-for-education.org/}{AI-for-Education Benchmarks}). This reinforces an important distinction: strong performance on abstract educational benchmarks does not necessarily translate into robust performance on situated, interactive instructional tasks. As a result, benchmark scores alone may substantially overestimate model readiness for deployment in discourse-sensitive educational analytics. 

\subsection{Prompting Significantly Improves Performance for Some Models}

Overall, adding examples to the prompt tended to improve performance, which is consistent with previous research \cite{Liu2025_GPT4Better}. Claude 4.5 Opus showed the most dramatic improvement, rising from $\kappa=0.48$ to $\kappa=0.58$, while Gemini 2.5 Pro improved from $\kappa=0.38$ to $\kappa=0.57$. This suggests that for models with sufficient context window and reasoning capacity, the inclusion of diverse, representative examples is a highly effective strategy for unlocking latent performance without the need for parameter updates. However, this benefit was not uniform as other models -- GPT-5 and o5 -- saw less and uneven benefit, indicating that the ability to leverage in-context learning for complex discourse is itself a differentiator of model quality.

\subsection{Implications for use of \textit{Out-of-the-Box} AI models for Educational Purposes}

Although many researchers and education technology developers have found successful institutional applications of AI, ranging from problem, hints, and feedback generation \cite{Meyer_2024} to AI-assisted human tutoring \cite{Wang2024_TutorCoPilot}, to tutor chat bots \cite{Labadze2023_REVIEWLitAIed}, many of these required fastidious research and practices to produce positive results. Given that students report using the LLMs directly through chat interfaces (e.g., ChatGPT), benchmarking the \textit{out-of-the-box} models is equally important to understand how these models can be optimized for educational tasks. Our results suggest that these models may not always process educational discourse in the same way that experts in instruction understand it. This may have downstream implications for how these models engage with students. 

This finding is contrasted by the models' performances on AI-for-Education's pedagogical benchmark task \cite{Lelièvre_2025}, which assesses these models using pedagogical tests similar to ones used in teacher education, with multiple choice questions on pedagogy. Thus, there is a need for further research on the conditions under which LLMs can be used for tasks related to educational discourse and instructors' behaviors.

\subsection{Implications for AI-Assisted Annotation}

Our findings provide important boundary conditions for the role of foundation models in educational annotation workflows. While prior studies demonstrate that LLMs can approach or even match human reliability under carefully designed pipelines \cite{He2024_GPT4Pipeline,Liu2025_GPT4Better}, those systems rely on structured verification, iterative refinement, or hybrid human–AI collaboration. In contrast, the present study isolates baseline competence and shows that unscaffolded foundation models are not yet sufficiently reliable to replace human coders for complex instructional constructs.

At the same time, the moderate performance observed across models indicates that foundation models exhibit latent sensitivity to certain pedagogical features. This suggests that future gains may come from principled hybrid systems that combine model suggestions with human verification, active learning, and error-sensitive auditing \cite{Wang2024_HumanLLMVerify,Nahum2025_LabelErrors} rather than from prompt engineering alone. Importantly, our findings caution against assuming that larger models or more examples will automatically reduce labeling error in discourse-rich domains.

\subsection{Limitations and Future Directions}

Several limitations should be noted. First, this study focused on a single well-validated discourse framework within mathematics instruction. While \textit{Talk Moves} provides a demanding test case, future work should examine whether similar performance baselines appear across other discourse taxonomies, subject areas, and interaction formats. Second, we evaluated only foundation models without retrieval, fine-tuning, or multi-stage reasoning pipelines. Although this was intentional for estimating baseline performance, hybrid systems may achieve substantially higher reliability in practice. Finally, while we examined six leading models, the rapid pace of model development means that absolute performance levels may shift. Nevertheless, the construct-specific patterns and prompt sensitivity observed here likely reflect more stable qualitative limitations.

Future research should pursue three complementary directions: (1) controlled ablation studies isolating which aspects of discourse most systematically degrade model performance, (2) hybrid human–AI workflows that explicitly target socially grounded inferential gaps, and (3) benchmark designs that distinguish between linguistic pattern recognition and genuinely interactional pedagogical reasoning.

\subsection{Conclusion}

Overall, this study provides the first systematic estimation of performance baselines for foundation models on a core instructional discourse task. While current models exhibit variability, we demonstrate that state-of-the-art foundation models (specifically Claude 4.5 Opus) can achieve moderate reliability ($\kappa \approx 0.58$) when supported by robust, example-driven prompts, although models tend to have high false positive rates. These results underscore that while "out-of-the-box" performance may be modest, prompt engineering---specifically the inclusion of diverse examples without cognitive overload---can effectively unlock the latent pedagogical capabilities of large foundation models. Future work should focus on closing the remaining gap to human experts through hybrid workflows and targeted fine-tuning.

\begin{credits}
\subsubsection{\ackname}

\end{credits}

\newpage
\section*{Appendices}
\appendix

\section{Per-Code Performance Details}
\label{App_precode}
The following tables present comprehensive per-code performance breakdowns for each model. Tables show Precision, Recall, F1, and Cohen's $\kappa$ for each talk move category across all three prompt strategies. Metrics are calculated using binary classification (target code vs. all other codes) to assess each model's ability to identify specific pedagogical moves.

\begin{table}[H]
\centering
\caption{Per-Code Performance for Claude 4.5 Opus.} 
\label{tab:per_code_claude_4_5_opus}
\begin{tabular}{p{3.5cm}>{ \centering \arraybackslash}p{2.0cm}>{ \centering \arraybackslash}p{2.0cm}>{ \centering \arraybackslash}p{2.0cm}>{ \centering \arraybackslash}p{2.0cm}}
  \toprule
Code & Precision & Recall & F1 & $\kappa$ \\ 
  \midrule \multicolumn{5}{l}{\textbf{Zero Shot}} \\ 
Keep Together & 0.35 & \textit{0.13} & \textit{0.19} & \textit{0.13} \\ 
  Relate to Ideas & 0.54 & 0.66 & 0.59 & 0.59 \\ 
  Restating & 0.51 & 0.88 & 0.65 & \textbf{0.64} \\ 
  Revoicing & 0.27 & 0.53 & 0.35 & 0.33 \\ 
  Press Accuracy & \textbf{0.72} & 0.62 & \textbf{0.66} & 0.62 \\ 
  Press Reasoning & \textit{0.20} & \textbf{0.96} & 0.33 & 0.32 \\ 
   \addlinespace[0.2em] \multicolumn{5}{l}{\textbf{One Shot}} \\ 
Keep Together & 0.36 & \textit{0.11} & \textit{0.17} & \textit{0.11} \\ 
  Relate to Ideas & 0.64 & 0.67 & 0.65 & 0.65 \\ 
  Restating & 0.54 & 0.89 & \textbf{0.67} & \textbf{0.67} \\ 
  Revoicing & 0.28 & 0.56 & 0.37 & 0.35 \\ 
  Press Accuracy & \textbf{0.74} & 0.61 & \textbf{0.67} & 0.63 \\ 
  Press Reasoning & \textit{0.20} & \textbf{0.98} & 0.34 & 0.32 \\ 
   \addlinespace[0.2em] \multicolumn{5}{l}{\textbf{Few Shot (Three Examples)}} \\ 
Keep Together & 0.52 & \textit{0.24} & \textit{0.33} & \textit{0.26} \\ 
  Relate to Ideas & 0.56 & 0.65 & 0.61 & 0.60 \\ 
  Restating & 0.54 & 0.89 & 0.67 & \textbf{0.67} \\ 
  Revoicing & 0.31 & 0.58 & 0.40 & 0.38 \\ 
  Press Accuracy & \textbf{0.75} & 0.68 & \textbf{0.71} & \textbf{0.67} \\ 
  Press Reasoning & \textit{0.24} & \textbf{0.98} & 0.39 & 0.38 \\ 
   \addlinespace[0.2em] \multicolumn{5}{l}{\textbf{Few Shot (All Examples)}} \\ 
Keep Together & 0.54 & \textit{0.55} & 0.54 & 0.47 \\ 
  Relate to Ideas & 0.64 & 0.75 & 0.69 & 0.68 \\ 
  Restating & 0.58 & 0.93 & \textbf{0.72} & \textbf{0.71} \\ 
  Revoicing & \textit{0.23} & 0.60 & \textit{0.33} & \textit{0.31} \\ 
  Press Accuracy & \textbf{0.77} & 0.62 & 0.69 & 0.65 \\ 
  Press Reasoning & 0.24 & \textbf{0.98} & 0.38 & 0.37 \\ 
   \bottomrule
\end{tabular}
\end{table}

\begin{table}[H]
\centering
\caption{Per-Code Performance for Claude 4.5 Sonnet.} 
\label{tab:per_code_claude_4_5_sonnet}
\begin{tabular}{p{3.5cm}>{ \centering \arraybackslash}p{2.0cm}>{ \centering \arraybackslash}p{2.0cm}>{ \centering \arraybackslash}p{2.0cm}>{ \centering \arraybackslash}p{2.0cm}}
  \toprule
Code & Precision & Recall & F1 & $\kappa$ \\ 
  \midrule \multicolumn{5}{l}{\textbf{Zero Shot}} \\ 
Keep Together & 0.30 & \textit{0.08} & \textit{0.13} & \textit{0.07} \\ 
  Relate to Ideas & 0.54 & 0.49 & 0.51 & 0.51 \\ 
  Restating & 0.42 & 0.85 & \textbf{0.56} & \textbf{0.55} \\ 
  Revoicing & 0.25 & 0.47 & 0.33 & 0.31 \\ 
  Press Accuracy & \textbf{0.58} & 0.48 & 0.52 & 0.46 \\ 
  Press Reasoning & \textit{0.16} & \textbf{0.93} & 0.28 & 0.26 \\ 
   \addlinespace[0.2em] \multicolumn{5}{l}{\textbf{One Shot}} \\ 
Keep Together & 0.33 & \textit{0.07} & \textit{0.11} & \textit{0.07} \\ 
  Relate to Ideas & 0.67 & 0.44 & 0.53 & 0.53 \\ 
  Restating & 0.43 & 0.88 & 0.58 & \textbf{0.57} \\ 
  Revoicing & 0.25 & 0.47 & 0.32 & 0.30 \\ 
  Press Accuracy & \textbf{0.69} & 0.53 & \textbf{0.60} & 0.55 \\ 
  Press Reasoning & \textit{0.19} & \textbf{0.92} & 0.31 & 0.30 \\ 
   \addlinespace[0.2em] \multicolumn{5}{l}{\textbf{Few Shot (Three Examples)}} \\ 
Keep Together & 0.47 & \textit{0.19} & \textit{0.27} & \textit{0.21} \\ 
  Relate to Ideas & 0.60 & 0.47 & 0.53 & 0.52 \\ 
  Restating & 0.43 & 0.88 & 0.58 & 0.57 \\ 
  Revoicing & \textit{0.23} & 0.54 & 0.33 & 0.30 \\ 
  Press Accuracy & \textbf{0.73} & 0.63 & \textbf{0.68} & \textbf{0.63} \\ 
  Press Reasoning & \textit{0.23} & \textbf{0.94} & 0.37 & 0.36 \\ 
   \addlinespace[0.2em] \multicolumn{5}{l}{\textbf{Few Shot (All Examples)}} \\ 
Keep Together & 0.44 & 0.53 & 0.48 & 0.39 \\ 
  Relate to Ideas & 0.56 & \textit{0.49} & 0.52 & 0.52 \\ 
  Restating & 0.46 & \textbf{0.91} & 0.61 & 0.60 \\ 
  Revoicing & \textit{0.20} & 0.62 & \textit{0.30} & \textit{0.28} \\ 
  Press Accuracy & \textbf{0.73} & 0.62 & \textbf{0.67} & \textbf{0.63} \\ 
  Press Reasoning & 0.26 & 0.90 & 0.40 & 0.39 \\ 
   \bottomrule
\end{tabular}
\end{table}

\begin{table}[H]
\centering
\caption{Per-Code Performance for Gemini 2.5.} 
\label{tab:per_code_gemini_2_5}
\begin{tabular}{p{3.5cm}>{ \centering \arraybackslash}p{2.0cm}>{ \centering \arraybackslash}p{2.0cm}>{ \centering \arraybackslash}p{2.0cm}>{ \centering \arraybackslash}p{2.0cm}}
  \toprule
Code & Precision & Recall & F1 & $\kappa$ \\ 
  \midrule \multicolumn{5}{l}{\textbf{Zero Shot}} \\ 
Keep Together & 0.26 & \textit{0.09} & \textit{0.14} & \textit{0.07} \\ 
  Relate to Ideas & 0.58 & 0.49 & 0.53 & 0.52 \\ 
  Restating & 0.52 & 0.84 & \textbf{0.64} & \textbf{0.64} \\ 
  Revoicing & 0.29 & 0.49 & 0.36 & 0.34 \\ 
  Press Accuracy & \textbf{0.65} & 0.39 & 0.48 & 0.43 \\ 
  Press Reasoning & \textit{0.17} & \textbf{0.97} & 0.28 & 0.27 \\ 
   \addlinespace[0.2em] \multicolumn{5}{l}{\textbf{One Shot}} \\ 
Keep Together & 0.35 & \textit{0.07} & \textit{0.11} & \textit{0.07} \\ 
  Relate to Ideas & 0.67 & 0.45 & 0.54 & 0.53 \\ 
  Restating & 0.49 & 0.83 & \textbf{0.61} & \textbf{0.61} \\ 
  Revoicing & 0.31 & 0.50 & 0.38 & 0.36 \\ 
  Press Accuracy & \textbf{0.74} & 0.51 & \textbf{0.61} & 0.56 \\ 
  Press Reasoning & \textit{0.15} & \textbf{0.99} & 0.26 & 0.25 \\ 
   \addlinespace[0.2em] \multicolumn{5}{l}{\textbf{Few Shot (Three Examples)}} \\ 
Keep Together & 0.48 & \textit{0.18} & \textit{0.27} & \textit{0.21} \\ 
  Relate to Ideas & 0.60 & 0.53 & 0.56 & 0.56 \\ 
  Restating & 0.54 & 0.86 & 0.66 & \textbf{0.66} \\ 
  Revoicing & 0.27 & 0.60 & 0.37 & 0.35 \\ 
  Press Accuracy & \textbf{0.69} & 0.65 & \textbf{0.67} & 0.62 \\ 
  Press Reasoning & \textit{0.19} & \textbf{0.97} & 0.32 & 0.30 \\ 
   \addlinespace[0.2em] \multicolumn{5}{l}{\textbf{Few Shot (All Examples)}} \\ 
Keep Together & 0.48 & \textit{0.54} & 0.51 & 0.42 \\ 
  Relate to Ideas & 0.62 & 0.62 & 0.62 & 0.61 \\ 
  Restating & 0.53 & 0.84 & 0.65 & 0.65 \\ 
  Revoicing & 0.29 & 0.60 & \textit{0.39} & \textit{0.37} \\ 
  Press Accuracy & \textbf{0.70} & 0.71 & \textbf{0.70} & \textbf{0.66} \\ 
  Press Reasoning & \textit{0.26} & \textbf{0.96} & 0.41 & 0.40 \\ 
   \bottomrule
\end{tabular}
\end{table}

\begin{table}[H]
\centering
\caption{Per-Code Performance for Gemini 3 Pro.} 
\label{tab:per_code_gemini_3_pro}
\begin{tabular}{p{3.5cm}>{ \centering \arraybackslash}p{2.0cm}>{ \centering \arraybackslash}p{2.0cm}>{ \centering \arraybackslash}p{2.0cm}>{ \centering \arraybackslash}p{2.0cm}}
  \toprule
Code & Precision & Recall & F1 & $\kappa$ \\ 
  \midrule \multicolumn{5}{l}{\textbf{Zero Shot}} \\ 
Keep Together & 0.38 & \textit{0.17} & \textit{0.24} & \textit{0.17} \\ 
  Relate to Ideas & 0.57 & 0.52 & 0.55 & 0.54 \\ 
  Restating & 0.51 & 0.82 & 0.63 & \textbf{0.62} \\ 
  Revoicing & 0.32 & 0.47 & 0.38 & 0.36 \\ 
  Press Accuracy & \textbf{0.66} & 0.62 & \textbf{0.64} & 0.59 \\ 
  Press Reasoning & \textit{0.18} & \textbf{0.97} & 0.31 & 0.29 \\ 
   \addlinespace[0.2em] \multicolumn{5}{l}{\textbf{One Shot}} \\ 
Keep Together & 0.39 & \textit{0.10} & \textit{0.16} & \textit{0.11} \\ 
  Relate to Ideas & 0.61 & 0.55 & 0.58 & 0.57 \\ 
  Restating & 0.47 & 0.85 & 0.60 & 0.60 \\ 
  Revoicing & 0.28 & 0.42 & 0.34 & 0.32 \\ 
  Press Accuracy & \textbf{0.70} & 0.63 & \textbf{0.66} & \textbf{0.61} \\ 
  Press Reasoning & \textit{0.18} & \textbf{0.96} & 0.31 & 0.29 \\ 
   \addlinespace[0.2em] \multicolumn{5}{l}{\textbf{Few Shot (Three Examples)}} \\ 
Keep Together & 0.47 & \textit{0.26} & \textit{0.33} & \textit{0.26} \\ 
  Relate to Ideas & 0.53 & 0.56 & 0.55 & 0.54 \\ 
  Restating & 0.55 & 0.81 & 0.65 & \textbf{0.65} \\ 
  Revoicing & 0.32 & 0.57 & 0.41 & 0.39 \\ 
  Press Accuracy & \textbf{0.72} & 0.68 & \textbf{0.70} & \textbf{0.65} \\ 
  Press Reasoning & \textit{0.25} & \textbf{0.97} & 0.40 & 0.39 \\ 
   \addlinespace[0.2em] \multicolumn{5}{l}{\textbf{Few Shot (All Examples)}} \\ 
Keep Together & 0.40 & 0.59 & 0.48 & \textit{0.38} \\ 
  Relate to Ideas & 0.58 & 0.65 & 0.62 & 0.61 \\ 
  Restating & 0.53 & 0.80 & 0.64 & 0.63 \\ 
  Revoicing & 0.34 & \textit{0.53} & \textit{0.41} & 0.39 \\ 
  Press Accuracy & \textbf{0.73} & 0.67 & \textbf{0.70} & \textbf{0.66} \\ 
  Press Reasoning & \textit{0.27} & \textbf{0.90} & 0.42 & 0.41 \\ 
   \bottomrule
\end{tabular}
\end{table}

\begin{table}[H]
\centering
\caption{Per-Code Performance for GPT-5.} 
\label{tab:per_code_gpt_5}
\begin{tabular}{p{3.5cm}>{ \centering \arraybackslash}p{2.0cm}>{ \centering \arraybackslash}p{2.0cm}>{ \centering \arraybackslash}p{2.0cm}>{ \centering \arraybackslash}p{2.0cm}}
  \toprule
Code & Precision & Recall & F1 & $\kappa$ \\ 
  \midrule \multicolumn{5}{l}{\textbf{Zero Shot}} \\ 
Keep Together & 0.33 & \textit{0.11} & \textit{0.17} & \textit{0.11} \\ 
  Relate to Ideas & 0.56 & 0.67 & 0.61 & 0.60 \\ 
  Restating & 0.63 & 0.80 & \textbf{0.70} & \textbf{0.70} \\ 
  Revoicing & 0.32 & 0.58 & 0.41 & 0.39 \\ 
  Press Accuracy & \textbf{0.65} & 0.64 & 0.64 & 0.59 \\ 
  Press Reasoning & \textit{0.17} & \textbf{0.97} & 0.30 & 0.28 \\ 
   \addlinespace[0.2em] \multicolumn{5}{l}{\textbf{One Shot}} \\ 
Keep Together & 0.35 & \textit{0.06} & \textit{0.11} & \textit{0.07} \\ 
  Relate to Ideas & 0.57 & 0.67 & 0.62 & 0.61 \\ 
  Restating & 0.62 & 0.80 & \textbf{0.70} & \textbf{0.70} \\ 
  Revoicing & 0.32 & 0.57 & 0.41 & 0.39 \\ 
  Press Accuracy & \textbf{0.67} & 0.64 & 0.65 & 0.60 \\ 
  Press Reasoning & \textit{0.19} & \textbf{0.98} & 0.31 & 0.30 \\ 
   \addlinespace[0.2em] \multicolumn{5}{l}{\textbf{Few Shot (Three Examples)}} \\ 
Keep Together & 0.45 & \textit{0.21} & \textit{0.28} & \textit{0.22} \\ 
  Relate to Ideas & 0.55 & 0.69 & 0.61 & 0.60 \\ 
  Restating & 0.66 & 0.86 & \textbf{0.74} & \textbf{0.74} \\ 
  Revoicing & 0.31 & 0.63 & 0.41 & 0.39 \\ 
  Press Accuracy & \textbf{0.68} & 0.66 & 0.67 & 0.62 \\ 
  Press Reasoning & \textit{0.20} & \textbf{0.98} & 0.33 & 0.31 \\ 
   \addlinespace[0.2em] \multicolumn{5}{l}{\textbf{Few Shot (All Examples)}} \\ 
Keep Together & 0.35 & 0.60 & 0.44 & \textit{0.32} \\ 
  Relate to Ideas & 0.61 & 0.69 & 0.65 & 0.64 \\ 
  Restating & 0.65 & 0.81 & \textbf{0.72} & \textbf{0.72} \\ 
  Revoicing & 0.32 & \textit{0.58} & 0.41 & 0.39 \\ 
  Press Accuracy & \textbf{0.69} & 0.70 & 0.69 & 0.65 \\ 
  Press Reasoning & \textit{0.25} & \textbf{0.98} & \textit{0.40} & 0.39 \\ 
   \bottomrule
\end{tabular}
\end{table}

\begin{table}[H]
\centering
\caption{Per-Code Performance for o3.} 
\label{tab:per_code_o3}
\begin{tabular}{p{3.5cm}>{ \centering \arraybackslash}p{2.0cm}>{ \centering \arraybackslash}p{2.0cm}>{ \centering \arraybackslash}p{2.0cm}>{ \centering \arraybackslash}p{2.0cm}}
  \toprule
Code & Precision & Recall & F1 & $\kappa$ \\ 
  \midrule \multicolumn{5}{l}{\textbf{Zero Shot}} \\ 
Keep Together & 0.38 & \textit{0.25} & 0.30 & \textit{0.22} \\ 
  Relate to Ideas & 0.49 & 0.63 & 0.56 & 0.55 \\ 
  Restating & 0.53 & 0.79 & \textbf{0.63} & \textbf{0.63} \\ 
  Revoicing & 0.30 & 0.51 & 0.38 & 0.36 \\ 
  Press Accuracy & \textbf{0.63} & 0.56 & 0.59 & 0.54 \\ 
  Press Reasoning & \textit{0.15} & \textbf{0.98} & \textit{0.26} & 0.24 \\ 
   \addlinespace[0.2em] \multicolumn{5}{l}{\textbf{One Shot}} \\ 
Keep Together & 0.40 & \textit{0.18} & \textit{0.25} & \textit{0.18} \\ 
  Relate to Ideas & 0.54 & 0.64 & 0.59 & 0.58 \\ 
  Restating & 0.50 & 0.78 & \textbf{0.61} & \textbf{0.60} \\ 
  Revoicing & 0.29 & 0.46 & 0.35 & 0.34 \\ 
  Press Accuracy & \textbf{0.66} & 0.56 & \textbf{0.61} & 0.55 \\ 
  Press Reasoning & \textit{0.15} & \textbf{0.97} & \textit{0.25} & 0.24 \\ 
   \addlinespace[0.2em] \multicolumn{5}{l}{\textbf{Few Shot (Three Examples)}} \\ 
Keep Together & 0.41 & \textit{0.33} & 0.37 & 0.28 \\ 
  Relate to Ideas & 0.51 & 0.65 & 0.57 & 0.56 \\ 
  Restating & 0.46 & 0.80 & 0.59 & \textbf{0.58} \\ 
  Revoicing & 0.30 & 0.52 & 0.38 & 0.36 \\ 
  Press Accuracy & \textbf{0.64} & 0.57 & \textbf{0.60} & 0.55 \\ 
  Press Reasoning & \textit{0.15} & \textbf{1.00} & \textit{0.26} & \textit{0.25} \\ 
   \addlinespace[0.2em] \multicolumn{5}{l}{\textbf{Few Shot (All Examples)}} \\ 
Keep Together & 0.33 & 0.60 & 0.43 & 0.31 \\ 
  Relate to Ideas & 0.59 & 0.68 & \textbf{0.63} & \textbf{0.62} \\ 
  Restating & 0.46 & 0.75 & 0.57 & 0.57 \\ 
  Revoicing & 0.29 & \textit{0.48} & 0.36 & 0.34 \\ 
  Press Accuracy & \textbf{0.67} & 0.60 & \textbf{0.63} & 0.58 \\ 
  Press Reasoning & \textit{0.17} & \textbf{0.94} & \textit{0.28} & \textit{0.27} \\ 
   \bottomrule
\end{tabular}
\end{table}

\newpage
\section{Pairwise Significance Tests}
\label{App_significance}

\setlength\tabcolsep{12pt}
\begin{longtable}{llrc}
\caption{Pairwise Bootstrap Significance Test ($R=1,000$) Model Performance Differences ($\kappa$)} 
\label{tab:pairwise_significance} \\
\toprule
\textbf{Model} & \textbf{Comparison} & \textbf{Diff.} & \textbf{$p$} \\
\midrule
\endfirsthead
\toprule
\textbf{Model} & \textbf{Comparison} & \textbf{Diff.} & \textbf{$p$} \\
\midrule
\endhead
\bottomrule
\endlastfoot
\textbf{Gemini 3 Pro} & FS (3) vs. FS (All) & -0.010 & 0.252 \\
 & FS (3) vs. 1-Shot & \textbf{0.052} & \textbf{< .001} \\
 & FS (3) vs. 0-Shot & \textbf{0.054} & \textbf{< .001} \\
 & FS (All) vs. 1-Shot & \textbf{0.057} & \textbf{< .001} \\
 & FS (All) vs. 0-Shot & \textbf{0.063} & \textbf{< .001} \\
 & 1-Shot vs. 0-Shot & 0.003 & 0.558 \\
\addlinespace[0.5em]
\textbf{Gemini 2.5} & FS (3) vs. FS (All) & \textbf{-0.070} & \textbf{< .001} \\
 & FS (3) vs. 1-Shot & \textbf{0.053} & \textbf{< .001} \\
 & FS (3) vs. 0-Shot & \textbf{0.119} & \textbf{< .001} \\
 & FS (All) vs. 1-Shot & \textbf{0.123} & \textbf{< .001} \\
 & FS (All) vs. 0-Shot & \textbf{0.188} & \textbf{< .001} \\
 & 1-Shot vs. 0-Shot & \textbf{0.065} & \textbf{< .001} \\
\addlinespace[0.5em]
\textbf{GPT-5} & FS (3) vs. FS (All) & \textbf{0.016} & \textbf{0.020} \\
 & FS (3) vs. 1-Shot & \textbf{0.026} & \textbf{< .001} \\
 & FS (3) vs. 0-Shot & \textbf{0.035} & \textbf{< .001} \\
 & FS (All) vs. 1-Shot & 0.010 & 0.182 \\
 & FS (All) vs. 0-Shot & \textbf{0.019} & \textbf{0.006} \\
 & 1-Shot vs. 0-Shot & \textbf{0.009} & \textbf{0.044} \\
\addlinespace[0.5em]
\textbf{o3} & FS (3) vs. FS (All) & \textbf{0.026} & \textbf{< .001} \\
 & FS (3) vs. 1-Shot & 0.006 & 0.276 \\
 & FS (3) vs. 0-Shot & 0.014 & 0.098 \\
 & FS (All) vs. 1-Shot & \textbf{-0.021} & \textbf{0.006} \\
 & FS (All) vs. 0-Shot & -0.013 & 0.200 \\
 & 1-Shot vs. 0-Shot & 0.007 & 0.402 \\
\addlinespace[0.5em]
\textbf{Claude 4.5 Opus} & FS (3) vs. FS (All) & \textbf{-0.046} & \textbf{< .001} \\
 & FS (3) vs. 1-Shot & \textbf{0.056} & \textbf{< .001} \\
 & FS (3) vs. 0-Shot & \textbf{0.062} & \textbf{< .001} \\
 & FS (All) vs. 1-Shot & \textbf{0.101} & \textbf{< .001} \\
 & FS (All) vs. 0-Shot & \textbf{0.106} & \textbf{< .001} \\
 & 1-Shot vs. 0-Shot & 0.006 & 0.070 \\
\addlinespace[0.5em]
\textbf{Claude 4.5 Sonnet} & FS (3) vs. FS (All) & \textbf{-0.029} & \textbf{< .001} \\
 & FS (3) vs. 1-Shot & \textbf{0.053} & \textbf{< .001} \\
 & FS (3) vs. 0-Shot & \textbf{0.087} & \textbf{< .001} \\
 & FS (All) vs. 1-Shot & \textbf{0.081} & \textbf{< .001} \\
 & FS (All) vs. 0-Shot & \textbf{0.116} & \textbf{< .001} \\
 & 1-Shot vs. 0-Shot & \textbf{0.035} & \textbf{< .001} \\
\end{longtable}

\newpage
\section{Prompt Templates}

This appendix provides the complete prompt templates used in our experiments, presented as JSON structures. All prompts include the same system role, workflow instructions, and code definitions, but differ in the number of examples provided.

\subsection{zero-shot Prompt}

\begin{lstlisting}[language=json]
{
  "role": [
    "You are an expert educational discourse analyst. Your task is to label each teacher utterance (Speaker = T) with one Talk Move from the allowed moves list."
  ],
  "workflow": [
    "Read the dialogue carefully.",
    "For each teacher utterance (Speaker = T), assign exactly ONE Talk Move from the Allowed Moves list.",
    "If an utterance could fit multiple moves, choose the one that best represents the communicative function in context.",
    "If no Talk Move applies (e.g., the utterance is off-topic, evaluative, unclear, or unrelated to the lesson), leave the TalkMove field blank."
  ],
  "allowedMoves": [
    "Keeping Everyone Together",
    "Getting Students to Relate to Another's Ideas",
    "Restating",
    "Pressing for Accuracy",
    "Revoicing",
    "Pressing for Reasoning"
  ],
  "moveDefinitions": {
    "Keeping Everyone Together": "Teacher prompts students to be active listeners and orienting students to each other.",
    "Getting Students to Relate to Another's Ideas": "Teacher prompts students to react to what a classmate said.",
    "Restating": "Teacher repeats all or part of what a student said word for word.",
    "Pressing for Accuracy": "Teacher prompts students to make a mathematical contribution or use mathematical language.",
    "Revoicing": "Teacher Repeats what a student said but adding on or changing the wording.",
    "Pressing for Reasoning": "Teacher prompts students to explain, provide evidence, share their thinking behind a decision, or connect ideas or representations."
  },
  "clarifications": [
    "Use only the six Allowed Moves exactly as written---no synonyms or rephrasings.",
    "Multiple consecutive utterances can have the same Talk Move.",
    "Do not code student utterances or teacher talk that does not elicit or build on student contributions.",
    "Exclude evaluative remarks (e.g., 'Good job!'), off-task talk, or administrative directions."
  ],
  "required_output": {
    "format": "JSON",
    "envelope": "Respond with a single JSON object with a top-level field named 'records'.",
    "fields": {
      "ID": "integer, exactly copied from the input utterance - THIS IS REQUIRED for matching.",
      "Speaker": "string, exactly copied from the input utterance (usually 'T').",
      "Turn": "integer, exactly copied from the input utterance.",
      "TalkMove": "one of the six Allowed Moves as a string, or null if no move applies."
    },
    "example": {
      "records": [
        {
          "ID": 1902,
          "Speaker": "T",
          "Turn": 150,
          "TalkMove": "Pressing for Reasoning"
        },
        {
          "ID": 1905,
          "Speaker": "T",
          "Turn": 153,
          "TalkMove": null
        }
      ]
    }
  }
}
\end{lstlisting}

\subsection{one-shot Prompt}

The one-shot prompt extends zero-shot by adding one example per talk move in the \texttt{moveExamples} field:

\begin{lstlisting}[language=json]
{  "moveExamples": {
    "Keeping Everyone Together": [
      "What did she say?"
    ],
    "Getting Students to Relate to Another's Ideas": [
      "Do you agree or disagree with him?"
    ],
    "Restating": [
      "Four million two."
    ],
    "Pressing for Accuracy": [
      "What is 6 times 6?"
    ],
    "Revoicing": [
      "So instead of one flat edge, it had two."
    ],
    "Pressing for Reasoning": [
      "Can you explain why?"
    ]
  }}
\end{lstlisting}

\subsection{few-shot Prompt (Three Examples)}
The few-shot  (All Examples) prompt extends zero-shot by adding multiple examples per talk move:
\begin{lstlisting}[language=json]
 {   "moveExamples": {
        "Keeping Everyone Together": [
            "So x equals five dollars, right?", 
            "It's going to be 150, right?", 
            "Are you finished?"],
        "Getting Students to Relate to Another's Ideas": [
            "How do you feel about what they said?", 
            "Does anyone understand how she solved the problem?", 
            "Do you agree or disagree with him?"],
        "Restating": [
            "S: The same size and shape but it moves to a different position. \n T: It moves to a different position. (Restating)",
            "S: An exponent \n  T: Exponent. (Restating)", 
            "S: It's four million and then the two. \n T: Four million two. (Restating)"],
        "Pressing for Accuracy": [
            "What is the answer to number 2?", 
            "How did you solve it?", 
            "What does x stand for?"],
        "Revoicing": [
            "S: It had two. \n T: So instead of one flat edge, it had two. (Revoicing)", 
            "S: Oh, Company B. \n T: It's Company B because that's the one that charges you $2.00 per minute. (Revoicing)", 
            "S: I got 2X minus Y. \n T: It's not 2X. (Revoicing)", 
            "S: La respuesta es siete. \n T: The answer is seven. (Revoicing)"],
        "Pressing for Reasoning": [
            "Can you explain why?", 
            "How are these ideas connected?", 
            "Where do we see the x + 1 in the tiles?"]
      }}
\end{lstlisting}

\subsection{few-shot Prompt (All Examples)}

The few-shot prompt extends zero-shot by adding multiple examples per talk move:

\begin{lstlisting}[language=json]
{  "moveExamples": {
    "Keeping Everyone Together": [
      "So x equals five dollars, right?",
      "Seven?",
      "Can we say these answers are the same?",
      "Go ahead.",
      "Shelly?",
      "Can you repeat what you just said?",
      "Turn and talk to your partner.",
      "Raise your hand if you know the answer.",
      "You know what?",
      "It's going to be 150, right?",
      "I know the dot goes here, okay?",
      "Are you finished?",
      "Where are your questions one, two, and three?",
      "Can you read the problem out loud?",
      "Repeat after me: common denominator.",
      "What did she say?",
      "Did you hear what he said?",
      "Someone tell me what he just said about his equation."
    ],
    "Getting Students to Relate to Another's Ideas": [
      "How do you feel about what they said?",
      "Do you know what he was doing?",
      "Would you like to add on to what he said?",
      "Does anyone understand how she solved the problem?",
      "Why did your classmates think that it was 7/2?",
      "Do you agree or disagree with him?",
      "Who is right?",
      "Raise your hand if you agree.",
      "Did you come up with something different than her?",
      "Did anyone else have three?"
    ],
    "Restating": [
      "It moves to a different position.",
      "Exponent.",
      "Four million two."
    ],
    "Pressing for Accuracy": [
      "What is the answer to number 2?",
      "Can you give me an example of an ordered pair?",
      "What is 6 times 6?",
      "Can you write your answer on the board?",
      "Show me where 6 inches is on the ruler.",
      "How can you figure that out?",
      "How did you solve it?",
      "What would I write next?",
      "What is this called?",
      "What's the definition of a square?",
      "What's another word for that?",
      "What does the .5 mean here?",
      "What does x stand for?"
    ],
    "Revoicing": [
      "So instead of one flat edge, it had two.",
      "The exponent.",
      "It's Company B because that's the one that charges you 
       $2.00 per minute.",
      "It's not 2X.",
      "The answer is seven."
    ],
    "Pressing for Reasoning": [
      "Can you explain why?",
      "How did you know?",
      "What made you extend these lines?",
      "Why does this method work?",
      "What's the point/What's the purpose?",
      "How are these ideas connected?",
      "How does this apply to ...?",
      "Where do we see the x + 1 in the tiles?",
      "How do we see that slope in your table of ordered pairs?",
      "How would that equation look on a graph?",
      "What would a table look like if I were to show that in 
       a systematic way?"
    ]
  }}
\end{lstlisting}


\begin{thebibliography}{99}

\bibitem{Zhang2024_GenAIReview}
Zhang, X., Zhang, P., Shen, Y., Liu, M., Wang, Q., Gašević, D., Fan, Y.:
A systematic literature review of empirical research on applying generative artificial intelligence in education.
Frontiers of Digital Education \textbf{1}(3), 223--245 (2024). 
\doi{10.1007/s44366-024-0028-5}  

\bibitem{Lee_2024}
Lee, S. S., Moore, R. L.:
Harnessing Generative AI (GenAI) for Automated Feedback in Higher Education: A Systematic Review.
Online Learning \textbf{28}(3), (2024). 
\doi{10.24059/olj.v28i3.4593}

\bibitem{Meyer_2024}
Meyer, J., Jansen, T., Schiller, R., Liebenow, L.W., Steinbach, M., Horbach, A., Fleckenstein, J.:
Using LLMs to bring evidence-based feedback into the classroom: AI-generated feedback increases secondary students’ text revision, motivation, and positive emotions.
Computers and Education: Artificial Intelligence \textbf{6}, 100199 (2024). 
\doi{10.1016/j.caeai.2023.100199}  

\bibitem{Turos2025AIHomework}
Turós, M., Nagy, R., Szűts, Z.:
What percentage of secondary school students do their homework with the help of artificial intelligence? – A survey of attitudes towards artificial intelligence.
Computers and Education: Artificial Intelligence \textbf{8}, 100394 (2025).
\doi{10.1016/j.caeai.2025.100394}

\bibitem{Adams2024_AIhelp}
Adams, D., Chuah, K.M., Devadason, E., Azzis, M.S.A.:
From novice to navigator: Students’ academic help-seeking behaviour, readiness, and perceived usefulness of ChatGPT in learning.
Education and Information Technologies \textbf{29}(11), 13617--13634 (2024).
\doi{10.1007/s10639-023-12427-8}

\bibitem{Mohammad2025_useChatGPT}
Mohammad, N.M., Demers, M., McCubbin, E., Mitchell, J., Fulmer, S.:
How college students use ChatGPT.
Pedagogical Research \textbf{10}(4), em0250 (2025).
\doi{10.29333/pr/17428}

\bibitem{VanDenBerg_2023}
Van Den Berg, G. \& Du Plessis, E. ChatGPT and Generative AI: Possibilities for Its Contribution to Lesson Planning, Critical Thinking and Openness in Teacher Education. Education Sciences 13, 998 (2023).
\doi{10.58806/ijsshmr.2025.v4i7n02}

\bibitem{DelValle_2025}
Del Valle, J. M. Use of Generative AI in Writing Lesson Plans: The Case of English Pre-Service Teachers. IJSSHMR 04, (2025).
\doi{10.58806/ijsshmr.2025.v4i7n02}

\bibitem{Diliberti_2025}
Diliberti, M. K., Lake, R. J. \& Weiner, S. R. More Districts Are Training Teachers on Artificial Intelligence: Findings from the American School District Panel. (2025).
\doi{https://www.rand.org/pubs/research_reports/RRA956-31.html}


\bibitem{otero2025benchmark}
Otero, N., Druga, S., Lan, A.:
A Benchmark for Math Misconceptions: Bridging Gaps in Middle School Algebra with AI-Supported Instruction.
Discover Education \textbf{4}, 277 (2025).

\bibitem{yang2025mmtutorbench}
Yang, T., et al.:
MMTutorBench: The First Multimodal Benchmark for AI Math Tutoring.
Preprint (2025).
\doi{10.48550/arXiv.2510.23477}

\bibitem{macina2025mathtutorbench}
Macina, J., et al.:
MathTutorBench: A Benchmark for Measuring Open-Ended Pedagogical Capabilities of LLM Tutors.
Preprint (2025).
\doi{10.48550/arXiv.2502.18940}


\bibitem{Lelièvre_2025}
Lelièvre, M. et al. Benchmarking the Pedagogical Knowledge of Large Language Models. Preprint at https://doi.org/10.48550/arXiv.2506.18710 (2025).
\doi{//doi.org/10.48550/arXiv.2506.18710}


\bibitem{Yimam2014_WebAnno}
Yimam, S.M., Biemann, C., Eckart de Castilho, R., Gurevych, I.:
Automatic annotation suggestions and custom annotation layers in WebAnno.
In: Bontcheva, K., Zhu, J. (eds.)
Proceedings of the 52nd Annual Meeting of the Association for Computational Linguistics: System Demonstrations,
pp.~91--96. Association for Computational Linguistics, Baltimore (2014).
\doi{10.3115/v1/P14-5016}

\bibitem{Henkel2023_ComparativeJudgment}
Henkel, O., Hills, L.:
Leveraging human feedback to scale educational datasets: Combining crowdworkers and comparative judgement.
In: Proceedings of the Tenth ACM Conference on Learning @ Scale,
pp.~411--415. Association for Computing Machinery, Copenhagen (2023).
\doi{10.1145/3573051.3596198}

\bibitem{Zambrano2023_nCoderChatGPT}
Zambrano, A.F., et al.:
From nCoder to ChatGPT: From automated coding to refining human coding.
In: Arastoopour Irgens, G., Knight, S. (eds.)
Advances in Quantitative Ethnography,
pp.~470--485. Springer Nature Switzerland, Cham (2023).
\doi{10.1007/978-3-031-47014-1_32}

\bibitem{Barany2024_CodebookDev}
Barany, A., Liu, X., Zhang, J., Pankiewicz, M., Baker, R.S.:
ChatGPT for education research: Exploring the potential of large language models for qualitative codebook development.
In: Olney, A.M., Chounta, I.-A., Liu, Z., Santos, O.C., Bittencourt, I.I. (eds.)
Artificial Intelligence in Education,
pp.~134--149. Springer Nature Switzerland, Cham (2024).
\doi{10.1007/978-3-031-64299-9_10}

\bibitem{He2024_GPT4Pipeline}
He, Z., Huang, C.-Y., Ding, C.-K.C., Rohatgi, S., Huang, T.-H.K.:
If in a crowdsourced data annotation pipeline, a GPT-4.
In: Proceedings of the 2024 CHI Conference on Human Factors in Computing Systems,
pp.~1--25. Association for Computing Machinery, New York (2024).
\doi{10.1145/3613904.3642834}

\bibitem{Liu2024_PotentialLimits}
Liu, X., Zhang, J., Barany, A., Pankiewicz, M., Baker, R.S.:
Assessing the potential and limits of large language models in qualitative coding.
In: Kim, Y.J., Swiecki, Z. (eds.)
Advances in Quantitative Ethnography,
pp.~89--103. Springer Nature Switzerland, Cham (2024).
\doi{10.1007/978-3-031-76335-9_7}

\bibitem{Liu2025_GPT4Better}
Liu, X., et al.:
Qualitative coding with GPT-4: Where it works better.
Journal of Learning Analytics \textbf{12}(1), 169--185 (2025).
\doi{10.18608/jla.2025.8575}

\bibitem{Geathers2025_OSCEBenchmark} Geathers, J., Hicke, Y., Chan, C.E., Rajashekar, N., Young, S., Sewell, J., Cornes, S., Kizilcec, R.F., Shung, D.: Benchmarking Generative AI for Scoring Medical Student Interviews in Objective Structured Clinical Examinations (OSCEs). In: Cristea, A.I., Walker, E., Lu, Y., Santos, O.C., Isotani, S. (eds.) Artificial Intelligence in Education. AIED 2025. Lecture Notes in Computer Science, vol. 15879, pp. 231–245. Springer, Cham (2025). \doi{10.1007/978-3-031-98420-4_17}

\bibitem{Long2024_ClassroomDialogue}
Long, Y., Luo, H., Zhang, Y.:
Evaluating large language models in analysing classroom dialogue.
npj Science of Learning \textbf{9}, 60 (2024).
\doi{10.1038/s41539-024-00273-3}

\bibitem{Nahum2025_LabelErrors}
Nahum, O., Calderon, N., Keller, O., Szpektor, I., Reichart, R.:
Are LLMs better than reported? Detecting label errors and mitigating their effect on model performance.
arXiv preprint (2024).
\doi{10.48550/arXiv.2410.18889}

\bibitem{Wang2024_HumanLLMVerify}
Wang, X., Kim, H., Rahman, S., Mitra, K., Miao, Z.:
Human–LLM collaborative annotation through effective verification of LLM labels.
In: Proceedings of the 2024 CHI Conference on Human Factors in Computing Systems,
pp.~1--21. Association for Computing Machinery, New York (2024).
\doi{10.1145/3613904.3641960}

\bibitem{Bojic2025_LatentContent}
Bojić, L., Zagovora, O., Zelenkauskaite, A., Vuković, V., Čabarkapa, M., Veseljević Jerković, S., Jovančević, A.:
Comparing large language models and human annotators in latent content analysis of sentiment, political leaning, emotional intensity and sarcasm.
Scientific Reports \textbf{15}, 11477 (2025).
\doi{10.1038/s41598-025-96508-3}

\bibitem{MoreauPernet_2024}
Moreau-Pernet, B. et al. Classifying Tutor Discursive Moves at Scale in Mathematics Classrooms with Large Language Models. in Proceedings of the Eleventh ACM Conference on Learning @ Scale 361–365 (ACM, Atlanta GA USA, 2024). 
\doi{doi:10.1145/3657604.3664664}

\bibitem{He2025_PromptingDark}
He, Z., Naphade, S., Huang, T.-H.K.:
Prompting in the dark: Assessing human performance in prompt engineering for data labeling when gold labels are absent.
In: Proceedings of the 2025 CHI Conference on Human Factors in Computing Systems,
pp.~1--33. Association for Computing Machinery, New York (2025).
\doi{10.1145/3706598.3714319}

\bibitem{Michaels2008AccountableTalk}
Michaels, S., O'Connor, C., Resnick, L.\,B.:
Deliberative discourse idealized and realized: Accountable talk in the classroom and in civic life.
Studies in Philosophy and Education \textbf{27}(4), 283--297 (2008).
\doi{10.1007/s11217-007-9071-1}


\bibitem{Suresh_2022_TalkMoves}
Suresh, A., Jacobs, J., Harty, C., Perkoff, M., Martin, J.H., Sumner, T.:
The TalkMoves Dataset: K–12 mathematics lesson transcripts annotated for teacher and student discursive moves.
In: Calzolari, N., Béchet, F., Blache, P., et al. (eds.)
Proceedings of the Thirteenth Language Resources and Evaluation Conference (LREC 2022),
pp.~4654--4662. European Language Resources Association, Marseille (2022).
\doi{10.48550/arXiv.2204.09652}


\bibitem{Wang2024_TutorCoPilot}
Wang, R.E., Ribeiro, A.T., Robinson, C.D., Loeb, S., Demszky, D.:
Tutor CoPilot: A Human-AI Approach for Scaling Real-Time Expertise.
arXiv preprint \doi{arXiv:2410.03017} (2024).

\bibitem{Labadze2023_REVIEWLitAIed}
Labadze, L., et al.:
Role of AI chatbots in education: systematic literature review.
International Journal of Educational Technology in Higher Education (2023).


\bibitem{LandisKoch1977}
Landis, J.R. \& Koch, G.G.:
The measurement of observer agreement for categorical data.
Biometrics \textbf{33}(1), 159–174 (1977). 
\doi{10.2307/2529310}


\end{thebibliography}
\end{document}